\documentclass[11pt]{article}

\usepackage[preprint]{acl}

\usepackage{times}
\usepackage{latexsym}

\usepackage[T1]{fontenc}

\usepackage[utf8]{inputenc}

\usepackage{microtype}

\usepackage{inconsolata}

\usepackage{graphicx}

\title{Intelligence Requires Grounding But Not Embodiment}

\author{Marcus Ma, Shrikanth Narayanan \\
  University of Southern California}

\begin{document}
\maketitle

\begin{abstract}

Recent advances in LLMs have reignited scientific debate over whether embodiment is necessary for intelligence. We present the argument that intelligence requires grounding, a phenomenon entailed by embodiment, but not embodiment itself. We define intelligence as the possession of four properties---motivation, predictive ability, understanding of causality, and learning from experience---and argue that each can be achieved by a non-embodied, grounded agent. We use this to conclude that grounding, not embodiment, is necessary for intelligence. We then present a thought experiment of an intelligent LLM agent in a digital environment and address potential counterarguments.

\end{abstract}

\section{Introduction}

\noindent The relationship between agentic intelligence and embodiment heavily divides the field of artificial intelligence into two broad camps of computational functionalism and embodied cognition. Functionalists believe intelligence is complex information processing independent of substrates and referents, with roots dating to the physical symbol system hypothesis \cite{newell1976computer}. Recent advances in LLMs seemingly support this hypothesis based on LLMs' emergent reasoning abilities \cite{wei2022emergent} and impressive zero-shot performance \cite{brown2020language}.

Proponents of embodied cognition argue intelligent agents must interact with the real world. They argue embodiment is necessary to prescribe meaning to symbols that otherwise lack them, a distinction known as the ``Symbol Grounding Problem'' \cite{harnad_1990}. From this viewpoint, intelligence requires coexistence with the dynamics of real-world interaction \cite{brooks1991intelligence} and comprehension of the rules of physics \cite{varela1991embodied, barsalou2008grounded}.

Recently, a minority view reconciling both positions has emerged. This middle ground argues that virtual embodiment, where agents interact in well-defined physics simulations to build their own world models, can produce intelligent agents \cite{lecun2022path}. \citet{bisk2020experience} similarly argue language comprehension requires grounded multimodal embodiment but this embodiment can be virtual. We clarify and extend this position: while embodiment, both virtual and physical, is sufficient for intelligence, it is not strictly necessary for it. The critical requirement is \textbf{grounding}: the mechanism by which to prescribe externally consistent meaning to symbols. When combined with interaction, grounding provides an environment capable of producing intelligent agents.

\section{Definitions}
\label{sec:definitions}

\begin{itemize}
\item An agent is an entity with the ability to sense and interact continuously in an environment through a perception-action loop. Critically, agents need not be embodied. A digital agent operates over data and information algorithmically rather than in the physical world. It receives input signals and produces output signals which then influence future input signals.

\item Grounding is the mechanism for which arbitrary symbols gain meaning \cite{harnad_1990} through the assignment of consistent, causal values to symbols based on referents existing in a reality external to the symbols themselves. Grounding ties the rules and constraints of the external environment to the symbols that aim to represent it and is how language and all symbolic systems gain meaning \cite{bisk2020experience, bender_2020}.

\item Embodiment refers to spatial presence in the physical world. While an implementation of a computational agent requires substrate hardware such as computers, the agent itself is substrate-independent, as its behaviors are not tied to its physical representation but rather its internal logic. In this work, we refer to non-embodied environments as digital.

\item A consensus definition of intelligence is widely debated \cite{legg_2007}. We opt to define intelligence extensionally, i.e., by defining a superset of traits demonstrative of intelligence such that possession of every trait constitutes intelligence. In Appendix~\ref{sec:intelligence}, we examine five established definitions of intelligence and define intelligence broadly as the possession of four characteristics: motivation, signal prediction, the ability to understand causality, and learning from experience.
\end{itemize}

\subsection{Distinguishing Intelligence from \\ \hspace{2.0em} Human Intelligence}

It is easy to over-define what constitutes intelligence based on bias towards the human experience. Human intelligence covers affect, consciousness, and an ability to create new goals autonomously; however, these are human properties rather than intelligent properties. An uncaring computer algorithm that only does as instructed could be intelligent as long as it achieves its goals effectively.

\section{Outline of the Core Argument}

\noindent We posit any agent with motivation, predictive power, understanding of causality, and experiential learning is intelligent. We claim none of these capabilities explicitly require embodiment but do require grounding. We thus conclude embodiment is not necessary for intelligence, but grounding is. We demonstrate how grounding alone is sufficient for a digital agent to possess each property:

\begin{itemize}

\item \S\ref{sec:motivation}: Motivation requires knowledge of a desired future state and a current state. A pure symbolic manipulation system has no mechanism to assign values to states; value must be derived from grounding to referents in the agent's environment.

\item \S\ref{sec:prediction}: Grounding is not required for predictive power. LLMs have demonstrated the ability of an entirely self-supervised computational model to predict the complex signal of human language with incredible accuracy.

\item \S\ref{sec:causality}: Understanding of causality naturally arises from the ability to predict, perceive, and interact as the agent learns which of its actions result in changes in the environment. This interaction must take place in a grounded environment with consistent rules and constraints, but this environment need not be embodied.

\item \S\ref{sec:improvement}: Learning from experience has two requirements: the ability to assign values to environment states and utilization of past experience to understand how to act to yield higher value states. For the first, we reuse our argument for motivation: value assignment relies on pre-established knowledge from an external grounding source. For the second, we reference decades of reinforcement learning research demonstrating that computational agents can autonomously find optimal behavior policies through unsupervised interaction in well-defined environments.

\end{itemize}

\noindent We conclude that because none of these properties explicitly requires embodiment, and possession of these four properties constitutes intelligence, embodiment is not necessary for intelligence.

\section{Intelligence Requires Sufficient Motivation}
\label{sec:motivation}

A core component of intelligence is the ability to accomplish goals; motivation is the driving force that sets goals and measures progress on them. Goal-driven action requires additional information beyond just symbol manipulation, as it requires a relative ordering of patterns of perception as ``good'' or ``bad'' with regards to a given goal. A purely predictive model treats all symbols as elements of a set with no relative ordering between elements; it only understands the correlations and patterns between symbols. This information must be grounded to referents that set the implicit or explicit values of specific perceptions, with motivation as the directive to move towards higher value states.

\subsection{On Intelligence and Biological Motivation}

There are several prominent theories that view intelligence as a prerequisite for life and intelligence as rooted deeply into the motivation for survival. Somatic marker theory posits that rationality (and thus intelligence) necessitates physical and biological regulation via homeostasis \cite{damasio_1994}. Autopoiesis, a theory describing the nature of living systems as a collection of self-replicating and growing components \cite{mingers_1991}, similarly sees ``living as a process [and] a process of cognition'' \cite{maturana_1980}. This could easily lead one to assume that embodiment in the form of an living organism is necessary for intelligence. However, we cannot assume being an embodied organism is required for intelligence just because all existing intelligent agents are embodied organisms. At its core, survival (i.e., maintaining one's homeostatic state) is just a difficult and long-spanning goal in the constantly changing and novel environment of the physical world. The physical world is high-fidelity, information dense, dynamic, and interactable, but none of these properties are tied to its embodiment. The ``homeostatic imperative'' \cite{damasio_2018} is simply a difficult enough task that intelligence is necessary to achieve it. If we could formulate a sufficiently complex goal (likely via multiple objectives) in a sufficiently complex digital environment, we could provide a digital agent with enough resources to be intelligent. \citet{silver_2021} even contend the maximization of an adequately complex reward is the only requirement for intelligent behavior; we agree broadly with this assessment but clarify the reward must be grounded in real-world utility to be a prerequisite for intelligence.

\section{Prediction Does Not Need Grounding}
\label{sec:prediction}

Grounding or meaning-making is not strictly necessary for prediction of a signal, even one as complex as language. In practice, self-supervised next-token prediction is a highly convincing objective for mimicking human language production. The fact that modern LLMs are capable of passing the Turing test \cite{jones_2025} is a proof by existence that the purely statistical structure of language paired with the existence of unfathomably large amounts of human text is sufficient to masterfully predict language generations without grounding. Whether or not language models can be grounded by text alone is discussed in Appendix~\ref{sec:language_grounding}; however, in both viewpoints, embodiment is certainly not necessary. In fact, accurate prediction of any digital signal can be derived through ungrounded statistical modeling.

\section{Causality Needs Interactivity}
\label{sec:causality}

To understand causal relationships, an agent must be able to identify how new and unseen changes in its perception and action result in corresponding changes in the environment even when not fully perceived. This requires the construction of an internal state and representation of a world model. Using LLMs as coordinators of reasoning and pattern-matching, there have been several historical attempts to properly model causality, including with probabilistic heuristic modeling \cite{wong_2023} and through proposals of complex physics simulations \cite{lecun2022path}. The critical component for learning causality is interactivity, wherein agents receive grounded feedback from their actions and build upon this feedback to draw connections between causes and effects. Interactivity consists of the perception-action loop, where agents observe, plan, and act continuously. \citet{brooks1991intelligence} argue perception and action are the only components necessary for intelligence; representational modeling and other computational paradigms are only optimizations to make intelligence tractable.

Without interactivity, a program manipulating only symbols can mimic and predict future occurrences of symbol patterns but could not generalize its capabilities beyond what it is trained on. \citet{searle1980minds} presents the Chinese room argument, where a program with a Chinese dictionary could translate a language into Chinese without ever knowing anything about it. Interactivity solves this problem by enabling the agent to enact change in the real world and provide perceptory feedback enabling causal understanding. In the real world, self-taught tool-augmented LLMs such as Toolformer \cite{schick2023toolformer}, applications such as model context protocol \cite{sanikommu_2025}, and reasoning-action training loops \cite{yao_2023} empirically demonstrate that adding {\em interactivity} to LLMs significantly improves performance and grounds LLM inference with the abilities of its tools.

\section{Intelligence Learns from Experience}
\label{sec:improvement}

We define learning from experience as an agent’s capacity to autonomously modify its internal or external states based on prior experiences in pursuit of a goal, which could be quantified as maximization of utility over an expected time span. This also includes the abilities of creating new sensors, tools, or skills and recursively breaking down complex tasks into smaller, more achievable goals \cite{good_1965, bostrom_2014}. We argue that while learning from experience requires understanding of and interactions with an environment, that environment need not be physical. We state two requirements: a mechanism for valuation (distinguishing good states from bad ones, where a state's ``goodness'' is determined by external grounded feedback), and a mechanism for optimization (navigation towards better states).

\subsection{Valuation via Grounding} We reuse our argument from \S\ref{sec:motivation}: an agent can only differentiate between good and bad through grounding. For example, reinforcement learning from human feedback \cite{christiano_2017, ouyang_2022} aligns LLM generations with human preference of better and worse responses.

\subsection{Optimization} Decades of research in reinforcement learning have demonstrated computational agents can master complex domains through self-supervised trial and error in a responsive environment \cite{sutton_barto_2018}. There are already many examples of RL agents in non-embodied environments that achieve superhuman capabilities in games like chess and Go \cite{silver_2017}, predict protein structures accurately \cite{jumper2021}, and discover mathematic algorithms \cite{Fawzi2022, Romera-Paredes2024}. Exploratory works in language processing find LLMs can adaptively learn by generating their own data filtered against a grounded verifier \cite{zelikman_2022, huang_2022}. Learning from experience requires grounding to know how to improve but does not require embodiment. Valuation and optimization in a consistent environment is sufficient.

\section{What Could Non-Embodied Intelligence Look Like?}

We now envision a potential non-embodied intelligent agent. While not strictly necessary for intelligence, we assert language manipulation would be the most feasible way to create human-interpretable intelligent reasoning given that language is the most reflective and abundant signal of human perception and belief. We consider an advanced tool-augmented and stateful language model that traverses the internet, a dynamic and complex digital environment. This agent runs on a computer server where it can write, read, and edit files as a form of memory and execute and create code as a means to create its own tools.

This agent directly communicates with a cooperative human, who sets arbitrary goals for the agent via dialogue. The agent can ask the human questions and request the human perform actions on its behalf (i.e., bypass a website with bot detection), but all ideation must come from the agent. The human only follows direct instructions, such that the accomplishment of the goal must occur from the agent's environmental manipulation. We argue if implemented optimally, this hypothetical agent could complete arbitrarily complex tasks on par with the capability of the cooperative human.

We consider how this agent might accomplish one such complex task intuitively requiring intelligent behavior to complete in a self-directed manner: making a certain amount of money in a fixed time span\footnote{We assume both computation and energy resources are abundant, such that there is essentially no cost for inference.}. To do so, this agent could examine its set of tools and identify ways to sell work, such as code fixes and data processing. It could also create basic software utilities to sell as products, such as paywalled APIs and text scrapers. The agent would navigate the internet to identify websites where it could perform this work and the agent would contact its cooperative human to transfer money, perform online verification, and bypass bot detection. The agent would analyze which of its methods yields the highest average return over time and explore new methods or scale existing ones accordingly. A highly advanced language agent could reason about this autonomously, using its tools and human communication as grounding and alignment with goal outcomes. There is no need for embodiment to accomplish this task or for many other complex tasks that could be performed online.

\section{Counterarguments}

We respond to potential counterarguments with elaboration in Appendices~\ref{sec:computers_embodied}, \ref{sec:low_bandwidth} and \ref{sec:physical_intelligence}.

\paragraph{Computers are embodied.} Computers are Turing machines, meaning their computations are not influenced by their physical forms.

\paragraph{Digital environments are too low-bandwidth to simulate reality.} This is a practical constraint, not a theoretical one. We concede artificial intelligence will likely arise \textit{first} from embodiment, given the complexity of creating a simulated and consistent environment; however, it is possible without.

\paragraph{Embodiment is necessary for physical intuition.} This may be true, but physical intuition is not necessary for intelligence. Physical intuition is not a requirement for enacting change or accomplishing goals in a non-embodied environment.

\clearpage

\section{Limitations}
This work presents a theoretical argument for why intelligence does not require embodiment. We are limited by the interpretations of our definitions, stated in \S\ref{sec:definitions} and Appendix~\ref{sec:intelligence}. Different interpretations of the terms defined could result in different conclusions than the ones drawn in this work.

\bibliography{custom}

\onecolumn

\appendix

\clearpage

\section{Defining Intelligence}
\label{sec:intelligence}

Intelligence itself is an elusive concept to objectively define. Rather than base our argument on a single definition, we collect five established definitions of intelligence and identify four general properties such that an agent with these four properties is intelligent for each definition. These properties are motivation, the ability to predict future states from action, the ability to understand causality, and learning from experience. We now examine these established definitions of intelligence and argue how possession of these four properties satisfies them:

\begin{enumerate}
    \item  \textbf{"Intelligence measures an agent's ability to achieve goals in a wide range of environments." \cite{legg_2007}}

    Goal achievement in diverse and unseen domains can be accomplished by setting a goal (motivation), interacting and exploring the environment (prediction and causality), and learning which actions contribute towards that goal (learning from experience).
    
    \item \textbf{"Intelligence is the efficiency with which a learner turns their experience and priors into new skills." \cite{chollet_2019}}

    Learning from experience requires a stateful understanding of utility (motivation), reshaping priors (prediction), and knowledge of how action shapes experience (causality).
    
    \item \textbf{"[Intelligence is] a very general mental capability that... involves the ability to reason, plan, solve problems... and learn from experience." \cite{gottfredson_1997}}

    We assert problem solving and planning are goal-directed (i.e., motivated) applications of reasoning. Reasoning, particularly causal reasoning, is the ability to infer how actions and interventions change the future \cite{pearl_2009}, which requires mastery of world prediction and causal understanding.
    
    \item \textbf{"[Intelligence is the] ability to understand complex ideas, to adapt effectively to the environment, to learn from experience, to engage in various forms of reasoning, [and] to overcome obstacles." \cite{neisser_1996}}

    Overcoming obstacles can be achieved through sufficient motivation and reasoning ability, while the rest of these properties have been covered above.
    
    \item \textbf{"[Intelligence is] the aggregate or global capacity of the individual to act purposefully, to think rationally, and to deal effectively with his environment." \cite{wechsler_1958}}

    Acting purposefully requires motivation, thinking rationally requires reasoning, and masterful manipulation of the environment requires causality and learning from experience.
\end{enumerate}

We thus assert that possession of these four properties is sufficient for intelligence based on these five definitions. In this work, we argue how grounding is required for all properties except prediction, and that no definition above can be satisfied by prediction alone, so therefore grounding is required for intelligence.

\newpage

\section{Can Text Alone Ground Meaning?}
\label{sec:language_grounding}

Below, we discuss two main viewpoints on whether text-only language models can ground meaning from symbolic manipulation of language.

\begin{enumerate}
    \item The most commonly accepted viewpoint is that text-only language models fail the Symbol Grounding Problem \cite{harnad_1990}, perhaps articulated most famously by \citet{bender_2020}. They argue real-world meaning cannot be derived from manipulations of inherently meaningless text symbols. \citet{stoljar_2024} assert because LLMs are not rational (in part due to their lack of grounding) and thinkers are inherently rational, LLMs cannot think.
    \item Recently, given the rise of emergent meaning-making properties in LLMs such as robust internal representations \cite{goldstein_2024} and learned spatiotemporal relationships \cite{gurnee_2024}, opposing viewpoints have emerged. \citet{mollo_2025} assert text-only LLMs are capable of deriving meaning for referents extending beyond their linguistic representation, meaning that internal language model states could theoretically be consistently and causally mapped to real-world outputs; \citet{gubelmann_2024} similarly argues language models are capable of referentially grounding their outputs to world meaning without multi-sensory or embodied perception.
\end{enumerate}

\noindent Ultimately, whether text alone can or cannot ground meaning does not affect our core argument; grounding is still required for intelligence whether inherent to the text or based on external labels.

\newpage

\section{Counterargument: Computers Are Embodied}
\label{sec:computers_embodied}

While computers technically occupy space, their physical form is an implementation of the underlying computation rather than a property of the computation itself. This is a critical component of the Church-Turing Thesis \cite{turing1936computable}, where the substrate (i.e., the hardware) of a given computer can be swapped independently with another computer and the final computational result does not change. All modern computers are Turing machines, meaning there exists a hypothetical Universal Turing Machine that can perform the exact same calculations as any other possible Turing machine indepedent of the machine's physical presence.

\newpage

\section{Counterargument: Digital Environments are Low-Bandwidth}
\label{sec:low_bandwidth}

A common counterargument to the existence of real-world intelligence derived from language alone, or any symbol or simulated signal, is the nature of discrete symbols and continuous reality. At a high enough resolution, the real world contains more information than can be conveyed via symbolic representation \cite{dreyfus1992computers}. For language specifically, ``tacit knowledge'' \cite{polanyi1966tacit} refers to information stemming from physical experience that cannot be distilled into linguistic representation.
\\ 

\noindent We respond to these critiques threefold: first, we argue that digital environments can be continuous; second, we argue that various aspects of human-level perception can be discretized; and third, we argue that multimodal interaction can bridge the gap between the low-bandwidth signal of a single input and the high-bandwidth information represented in the real world.

\subsection{Continuity from Discrete Samples} Discrete digital systems are capable of representing continuous signals. The Nyquist-Shannon sampling theorem states that continuous signals can be reconstructed without loss from discrete samples given a sampling rate twice as high as the signal's highest frequency component \cite{shannon1949communication}. Additionally, neural networks are also able to approximate continuous functions to an arbitrary level of precision with a finite number of neurons as long as the precision is defined before training \cite{hornik1989multilayer}.

\subsection{Human Perception Can Effectively Be Discretized} \citet{vanrullen2003perceptual} claim human perception of reality is actually discrete. Neural communication arises from action potentials, or binary thresholds of energy that either spike or remain dormant \cite{hodgkin1952quantitative}. Additionally, sensory perception such as audio and vision have limits: after a certain threshold of resolution is achieved, humans cannot tell the difference between continuous sight and an extremely high-resolution image, based on the mechanistic sensitivity of cone photoreceptors in the eye \cite{curcio1990human, wandell1995foundations}, and after a certain sampling rate, human listeners cannot determine which audio is higher resolution \cite{meyer2007audibility}.

\subsection{Multimodal Synergy} Text by itself may face issues of tacit knowledge as a low-bandwidth signal; however, when paired with multimodal interaction, integrating text with other kinds of modalities allows construction of meaning based on relationships between these signals \cite{bisk2020experience}. This effectively raises the informational bandwidth that an environment can convey via high-dimensional interactions between modalities.

\newpage

\section{Counterargument: Embodiment is Necessary for Physical Intuition}
\label{sec:physical_intelligence}

Artificial intelligence currently suffers Moravec's paradox, the observation that computers excel at symbolic reasoning challenges such as chess and mathematics fairly easily but require intense computational resources to handle tasks that come very easily to humans such as grasping an object, socializing, and recognizing emotions \cite{moravec1988mind}. We respond to this in two ways: first, we posit physical intuition is simply not necessary for intelligence in general, i.e., it is possible to be intelligent without knowledge of or existence in the physical world. Second, we argue that physical intuition \textbf{can theoretically} be learned without embodiment, but that embodiment makes the problem significantly more tractable.

\subsection{Intelligence Does Not Need Physical Intuition}
Intelligence refers to mastery over one's environment (see Appendix~\ref{sec:intelligence}). If this environment is not embodied, then an agent in this environment does not need physical intuition to be considered intelligent.

\subsection{Physical Intuition Can Be Learned Without Embodiment}

Properties of the physical world can be inferred from digital representations such as images and audio. Recent works in machine learning have demonstrated that AI systems can learn physical concepts such as gravity purely from visual stimuli \cite{piloto2022intuitive, riochet2018intphys}. While learning gravitational intuition would likely require much less computation and resources with an embodied system, it is \textbf{possible} to observe this phenomenon from representational artifacts. We argue that embodiment enables computationally efficient ways of learning and feeling physical phenomena, but it is not strictly necessary to be able to model them. With enough data and compute, a physical intuition of the world could be learned. However, we concede that this is extremely inefficient compared to embodiment and that an embodied sensory experience may be the only tractable way to model physical intuition.

\end{document}